\documentclass[conference]{IEEEtran}
\IEEEoverridecommandlockouts
\usepackage{cite}
\usepackage{amsmath,amssymb,amsfonts}
\usepackage{algorithmic}
\usepackage{graphicx}
\usepackage{textcomp}
\usepackage{xcolor}

\usepackage{booktabs}
\usepackage[caption=false]{subfig}
\def\BibTeX{{\rm B\kern-.05em{\sc i\kern-.025em b}\kern-.08em
    T\kern-.1667em\lower.7ex\hbox{E}\kern-.125emX}}

\usepackage{url}
\usepackage[hidelinks]{hyperref}
\begin{document}

\title{Tell Me: An LLM-powered Mental Well-being Assistant with RAG, Synthetic Dialogue Generation, and Agentic Planning
}

\author{\IEEEauthorblockN{1\textsuperscript{st} Trishala Jayesh Ahalpara}
\IEEEauthorblockA{\textit{Fujitsu Research of America} \\
Santa Clara, USA \\
tahalpara@fujitsu.com}
}

\IEEEoverridecommandlockouts
\IEEEpubid{
  \parbox[t]{\columnwidth}{\vspace{8pt}\raggedright
  \copyright~2026 IEEE. Personal use of this material is permitted. Permission from IEEE must be obtained for all other uses, in any current or future media, including reprinting/republishing this material for advertising or promotional purposes, creating new collective works, for resale or redistribution to servers or lists, or reuse of any copyrighted component of this work in other works.}%
  \hspace{\columnsep}\makebox[\columnwidth]{}%
}

\maketitle

\begin{center}
\small
This paper has been accepted for publication in the Proceedings of the
2nd International Conference on Sustainability, Innovation, and Society
(ICSIS 2026). The final authenticated version will be available through
IEEE Xplore.
\end{center}

\begin{abstract}
We present \textit{Tell Me}, a mental well-being system that leverages recent advances in large language models to provide accessible, context-aware support for users and researchers. The system integrates three components: (i) a retrieval-augmented generation (RAG) assistant for personalized, knowledge-grounded dialogue; (ii) a synthetic client--therapist dialogue generator conditioned on client profiles to facilitate research on therapeutic language and data augmentation; and (iii) a Well-being AI crew, implemented with CrewAI, that produces weekly self-care plans and guided meditation audio. The system is designed as a reflective space for emotional processing rather than a substitute for professional therapy. It illustrates how conversational assistants can lower barriers to support, complement existing care, and broaden access to mental health resources. To address the shortage of confidential therapeutic data, we introduce synthetic client--therapist dialogue generation conditioned on client profiles. Finally, the planner demonstrates an innovative agentic workflow for dynamically adaptive, personalized self-care, bridging the limitations of static well-being tools. We describe the architecture, demonstrate its functionalities, and report evaluation of the RAG assistant in curated well-being scenarios using both automatic LLM-based judgments and a human-user study. This work highlights opportunities for interdisciplinary collaboration between NLP researchers and mental health professionals to advance responsible innovation in human--AI interaction for well-being.
\end{abstract}


\begin{figure*}[t]
    \centering
    \includegraphics[width=0.9\textwidth]{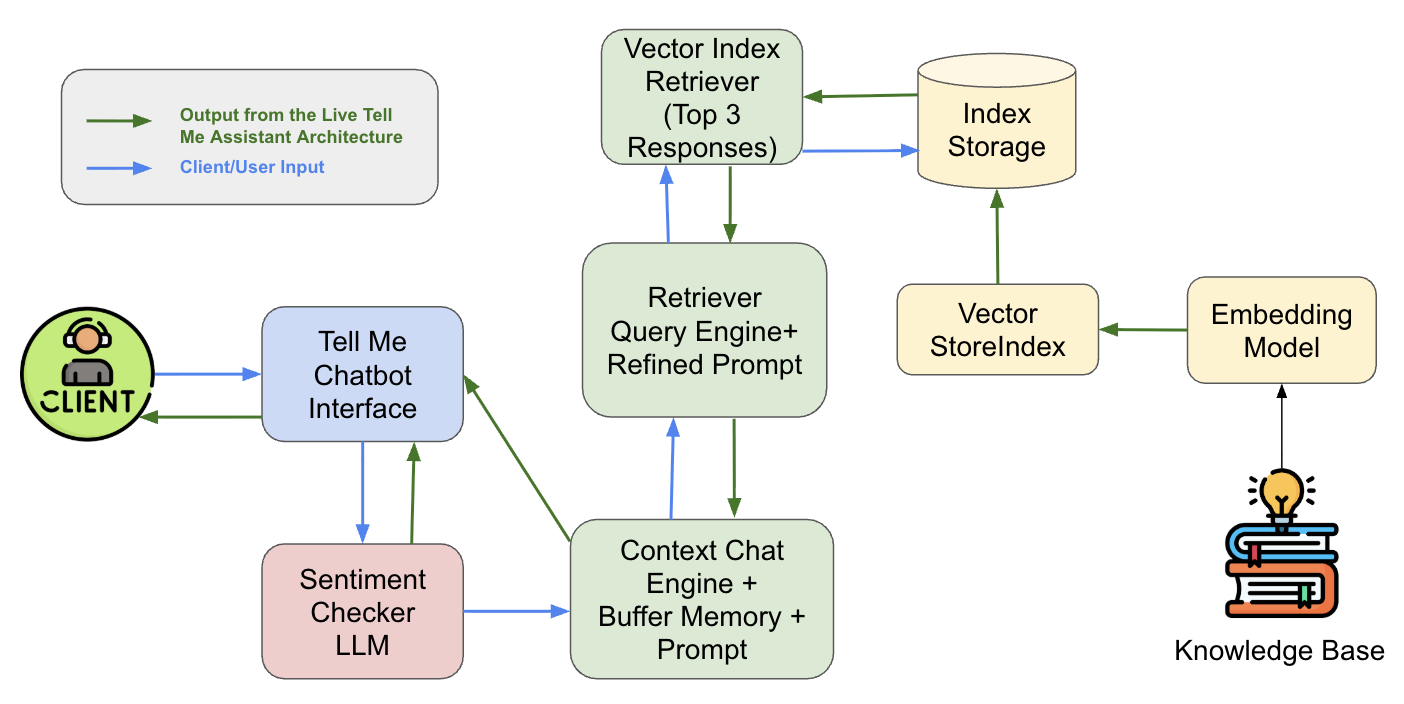}
    \caption{System architecture of the \textit{Tell Me Mental Well-being Assistant}, showing integration of RAG, synthetic dialogue generation, and agentic AI modules.}
    \label{fig:system-architecture}
\end{figure*}

\section{Introduction}

Recent years have seen rapid growth in the use of Large Language Models (LLMs) in mental well-being applications: from therapeutic chatbots for emotional support \cite{song2024typing}, to screening and psychotherapeutic interventions via smart/everyday devices \cite{nie2024caiti}, simulated client assessments \cite{wang2024clientcast}, and LLM evaluation frameworks \cite{chiu2024bolt}. Other efforts explore fine-tuning and prompt design for mental health \cite{yu2024fine_tuned_llm} and benchmarking LLMs against human peers in cognitive behavioral therapy (CBT) \cite{iftikhar2024therapy}. Despite this momentum, most systems remain limited in scope and face three persistent gaps: (1) responses often lack sufficient context, leading to generic or shallow interactions; (2) access to real therapeutic dialogues is restricted by confidentiality, limiting dataset availability and reproducibility; and (3) many well-being tools are static, offering little adaptation to individual needs or evolving emotional states.

We introduce \textit{Tell Me}, a demo system designed to close these gaps and serve as a lightweight, extensible testbed for responsible LLM applications in mental well-being. \textit{Tell Me} is available as an open-source prototype via Hugging Face Spaces\footnote{Hugging Face Demo: \url{https://huggingface.co/spaces/trystine/Tell_Me}}
and GitHub\footnote{GitHub: \url{https://github.com/trystine/Tell_Me_Mental_Wellbeing_System}}.

\vspace{0.5em}
\noindent Specifically, our contributions are:
\begin{itemize}
    \item \textbf{A context-sensitive RAG assistant} that enables reflective, knowledge-grounded dialogue to support emotional processing. We benchmarked nine LLMs across ten situational well-being prompts using an LLM-as-a-judge framework, and validated the best-performing model through a blind human study comparing RAG vs.\ non-RAG outputs.
  \item \textbf{A synthetic client--therapist dialogue generator} based on user profiles, enabling safe, customizable data augmentation without compromising patient confidentiality. We provide a case study demonstrating its utility in generating realistic, profile-driven therapeutic exchanges.
    \item \textbf{An agentic well-being planner (CrewAI)} that translates conversations into dynamically adaptive self-care practices. An accompanying case study illustrates how this system extends support into actionable steps, such as weekly planning and guided meditation.
\end{itemize}

This work is intended as a reflective support space rather than a substitute for professional therapy. However, conversational assistants can help reduce barriers to support and complement existing care. Previous studies show that many clients withhold sensitive information in therapy due to fear of judgment~\cite{hill2015client, khawaja2023chatbots}, and recent reviews highlight that disclosure and trust remain persistent challenges for digital mental health tools~\cite{mayor2025chatbots}. By designing systems with empathy and safeguards, we aim to support responsible innovation in human--AI interaction for well-being.

\section{Related Work}

Conversational agents for mental health have a long history, beginning with \textit{ELIZA} \cite{weizenbaum1966eliza} and expanding to commercial platforms such as Woebot \cite{fitzpatrick2017delivering}, Wysa \cite{inkster2018wysa}, and Replika \cite{skjuve2021replika}, which demonstrated scalability and measurable impact in controlled settings. However, these systems remain closed-source and nonreproducible, limiting their value for research.

With the rise of LLMs, conversational systems have become more adaptive and empathetic. Recent studies explore emotional support, simulated client assessments, LLM therapist evaluation, and CBT-style comparisons with human peers \cite{song2024typing,wang2024clientcast,chiu2024bolt,iftikhar2024therapy}, while others investigate fine-tuning and prompt design for mental health chatbots \cite{yu2024fine_tuned_llm}. These advances highlight potential, but also underscore persistent concerns about reliability, safety, and confidentiality. Access to authentic therapeutic dialogues is restricted, motivating profile-conditioned synthetic dialogues as a safer alternative \cite{wang2024clientcast}.

Retrieval-augmented generation (RAG) improves factuality and relevance in dialogue \cite{lewis2020rag,shuster2021retrieval}, with more recent systems such as BlenderBot 3 \cite{shuster2022blenderbot3}, INFO-RAG \cite{xu2024inforag}, and R2AG \cite{ye2024r2ag} extending retrieval for open-domain conversation, safety, and long-term grounding. Our system differs by applying RAG specifically to mental well-being, where the goal is not only factuality, but also reflective, empathetic, and context-sensitive dialogue. This positions \textit{Tell Me} as a demonstration of how retrieval grounding can enhance responsible interaction in emotionally sensitive domains.

Parallel efforts in healthcare explore agentic AI, including multiagent dialogue support for clinicians \cite{kampman2024dual} and psychiatric interview assistants \cite{bi2025magi}. These works primarily target diagnostic or clinician-facing use cases. In contrast, our system employs CrewAI-based orchestration to extend end-user support to dynamic planning and guided meditation, addressing the limitations of static wellness tools.

\textit{Tell Me} is the first open demo to showcase three independent but complementary modules: a context-aware RAG assistant, a synthetic dialogue generator addressing data scarcity, and an agentic planner for adaptive self-care. Presented side by side, they highlight different pathways for advancing safe and customizable well-being support.

\begin{figure*}[t]
    \centering
    \subfloat[Simulate a Conversation module.]{
        \includegraphics[width=0.485\textwidth]{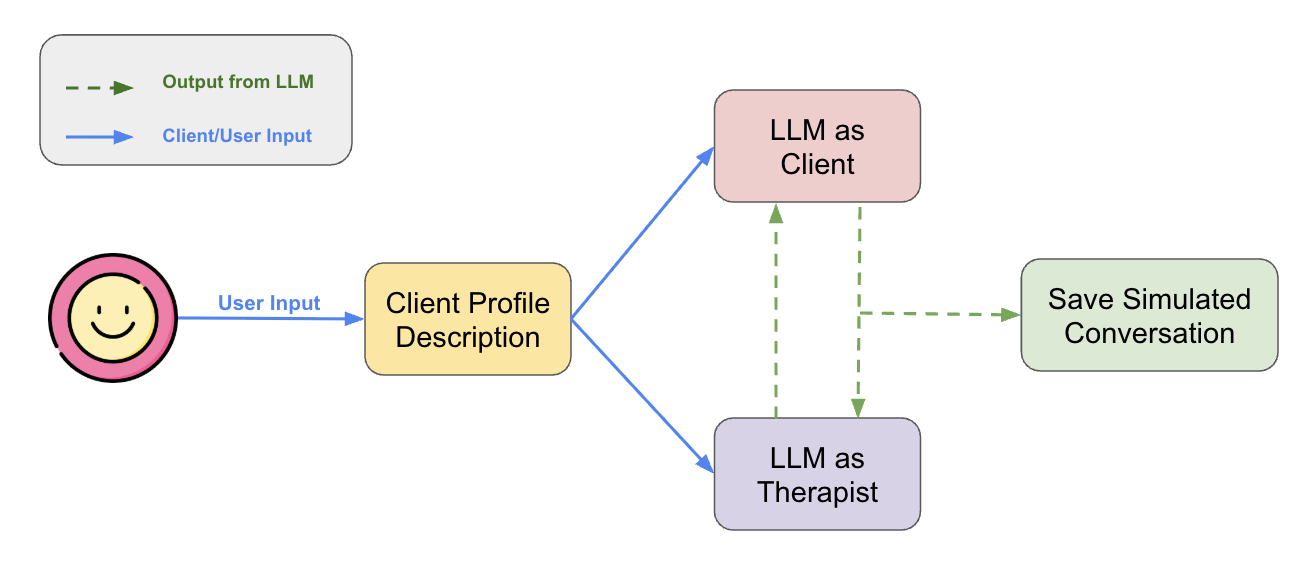}
        \label{fig:diagram2}
    }
    \hfil 
    \subfloat[Well-being Planner with CrewAI agents.]{
        \includegraphics[width=0.485\textwidth]{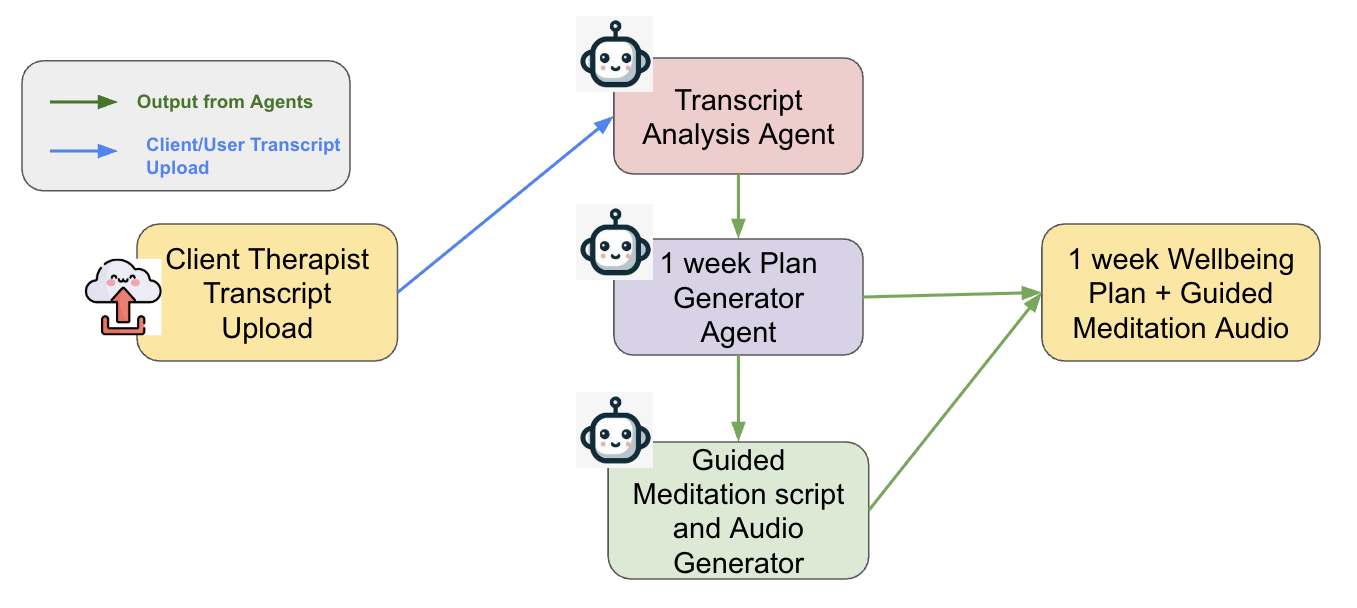}
        \label{fig:diagram3}
    }
    \caption{System architecture components of the \textit{Tell Me Mental Well-being Assistant}: (a) synthetic conversation simulation; (b) CrewAI-based well-being planner.}
    \label{fig:system-modules}
\end{figure*}

\section{System Overview}
\label{sec:system-overview}

\textit{Tell Me} comprises three modular components: (i) a RAG-based assistant for context-aware reflective dialogue, (ii) a synthetic conversation generator for data augmentation, and (iii) a CrewAI-based planner for actionable self-care orchestration. The system architecture is unified via a lightweight Streamlit front-end \cite{streamlit2022}, which manages global application state, coordinates user configuration keys, and dynamically routes inference calls. The unified interface allows users to switch seamlessly between cloud-based LLMs (e.g., OpenAI GPT-4o, Anthropic Claude 3.7 Sonnet) and local instances via Ollama. End-users can engage with the assistant and planner for reflective support, while researchers can generate safe synthetic transcripts for experimentation. 

\subsection{RAG-Based Assistant}
The RAG-based assistant forms the core client-facing module, providing context-aware responses grounded in a curated knowledge base. Unlike baseline LLMs that often provide generic or verbose advice, this assistant integrates embedding, retrieval, and context management to maintain conversational history while offering relevant and grounded responses. 

To ensure safe interaction, the system employs an LLM-driven guardrail mechanism (\texttt{Sentiment\_chain} via LangChain) as a pre-processing stage. User inputs are evaluated for high-risk semantic signals, such as expressions of self-harm or harmful intent. When such signals are detected, the pipeline suppresses free-form response generation and activates a deterministic crisis-routing mechanism. This approach constrains the conversational agent’s behavior, enabling \textit{Tell Me} to emulate therapist-style interactions while maintaining adherence to safety and ethical guidelines.

\subsubsection{Data and Knowledge Base}
To address confidentiality constraints, we adopt a Context–Response schema where context represents a client’s input and response represents a therapist-like reply. For our prototype, we use two open-source datasets—Counsel Chat \cite{demasi2019counselchat} and Mental Health Counseling Conversations \cite{amod2023mentalhealthconvos}—reformatted into this schema to serve as the retrieval base.

\subsubsection{Embedding, Retrieval, and Context Management}

For document vectorization and semantic search, we utilize the \texttt{BAAI/bge-small-en-v1.5} model \cite{baai_bge_small_en_v1_5}. This model was selected over larger alternatives for three primary reasons: (i) \textit{Performance:} it ranks highly on the Massive Text Embedding Benchmark (MTEB) among lightweight models for retrieval-centric tasks; (ii) \textit{Inference Efficiency:} its 384-dimensional embedding space reduces latency and memory overhead; and (iii) \textit{Improved Similarity Calibration:} the v1.5 iteration refines similarity score distributions to mitigate embedding clustering effects.

Documents are segmented using a \texttt{TokenTextSplitter} (chunk size = 128 tokens, overlap = 20) and stored in a persistent LlamaIndex \texttt{VectorStoreIndex}. During inference, a \texttt{VectorIndexRetriever} selects the top-3 most relevant context-response pairs based on cosine similarity. These candidates are processed by a \texttt{RetrieverQueryEngine}, which is further wrapped by a \texttt{ContextChatEngine} to guide the downstream LLM toward maintaining a compassionate, non-clinical therapeutic tone.

Multi-turn conversational coherence is maintained using a \texttt{ChatMemoryBuffer} with a 3,000-token limit. Historical dialogue is appended directly to the synthesis prompt, ensuring contextual continuity while minimizing conversational drift.

\subsubsection{Modes of Operation}
The application state manager supports two distinct execution flows: (i) \textit{Public Mode}, where users interact directly with the system as a reflective chatbot with toggleable RAG functionality, and (ii) \textit{Study Mode}, a blinded A/B testing environment used in our human evaluation. Study mode implements access gates, sequential routing (RAG vs. Non-RAG), and integrated Likert-scale data collection.

\subsection{Simulate a Conversation}
To mitigate the shortage of confidential therapeutic data, this module generates synthetic dialogues by role-playing both client and therapist. Researchers define a client profile (e.g., demographics, concerns, history), and the system produces transcripts conditioned on these attributes. 

\begin{sloppypar}\raggedright
Architecturally, the simulation is implemented via LangChain using two isolated \texttt{LLMChain} objects equipped with \texttt{Conversation Buffer Memory}. The \textit{Client Chain} is initialized with a system prompt containing the user-defined persona, while the \textit{Therapist Chain} is instructed to use reflective listening techniques. The application logic seeds an initial therapist greeting and then passes the chat history iteratively between the two models in a closed loop. The final multi-turn transcript is collected and exposed to the user interface as a downloadable text file.\par
\end{sloppypar}

\subsection{Well-being Planner}
The planner extends the dialogue into actionable support using CrewAI \cite{crewai}. It coordinates three distinct agents powered by GPT-4o (\textit{temperature = 0.7}): (i) a \textit{Transcript Analyzer} that extracts emotions and concerns, (ii) a \textit{Plan Generator} that creates a formatted seven-day routine, and (iii) a \textit{Meditation Generator} that scripts a guided audio session. 

To ensure deterministic execution, agent coordination relies on sequential task dependencies. The output of the Transcript Analyzer acts as the strict contextual input for the Plan Generator. Once the text-based weekly routine and meditation scripts are generated, the raw meditation text is piped asynchronously through a Text-to-Speech (TTS) integration wrapper (supporting providers like Edge Neural TTS, ElevenLabs, or Coqui). The generated audio byte stream is compiled into an MP3 file for the user to download alongside their weekly plan.

\subsection{Availability}
The live interactive demonstration is publicly available on Hugging Face Spaces \cite{tellme_space}, and the source code repository has been open-sourced on GitHub \cite{tellme_github}. To prioritize end-user data privacy and limit structural runtime maintenance costs, the deployment architecture operates an ephemeral key-passing model requiring users to supply personal API credentials.

\section{RAG Assistant Evaluation}

We evaluated the RAG-based assistant; the other modules (synthetic dialogue generator and planner) are presented as functional demonstrations with example case studies (Section~\ref{sec:case-study-synthetic}, Section~\ref{sec:case-study-agentic}) . Our protocol has two phases: (i) an automatic \emph{LLM-as-a-judge} benchmark on curated prompts, and (ii) a human study comparing the best model in RAG vs. non-RAG settings.

\subsection{Benchmark Setup}

\subsubsection{Prompts}
We design 10 scenario prompts covering common well-being themes: loneliness \& social comparison, anxiety, depression, grief, low self-esteem, abuse, relationships, family dynamics, fear, and addiction. These reflect issues frequently addressed in therapy: loneliness and lack of support are known mental health risk factors \cite{cdc_loneliness_2024}, anxiety and depression are the most common targets of digital interventions \cite{casu2024chatbots}, family dynamics are central in therapy \cite{nichols2020family}, and addiction is a major public health focus \cite{samhsa2022addiction}.

\subsubsection{Candidate models}
We evaluate nine representative models spanning frontier LLMs and well-being--tuned systems~\cite{openai2024gpt4o,touvron2023llama,jiang2023mistral,gemma2024,anthropic2025claude,microsoft2024phi4}: GPT-4o, LLaMA-3, Mistral-7B, Gemma-3, Claude 3.7 Sonnet, and Phi-4 Mini 3.8B, plus three Ollama-based models.\footnote{Ollama models: \texttt{vitorcalvi/mentallama2:latest}, \texttt{wmb/llamasupport}, \texttt{ALIENTELLIGENCE/mentalwellness}.} Each model answers all prompts. The top performer advances to RAG vs. non-RAG ablations (Section~\ref{sec:human-eval}).

\subsection{Retrieval Evaluation}

To evaluate the retrieval component independently of downstream generation, we conducted a retrieval-only assessment of the RAG pipeline using synthetic query-document relevance pairs. We randomly sampled 30 counseling interactions from the combined CounselChat and Mental Health Counseling Conversations datasets and indexed them using the \texttt{BAAI/bge-small-en-v1.5} embedding model. Each interaction merges the client context and therapist response into a single retrieval document.

Due to the absence of a human-labeled retrieval benchmark, we adopt a synthetic ground-truth protocol implemented via LlamaIndex. A local \texttt{LLaMA-3} model (served through Ollama) generates one representative query per document, forming paired query-document relevance sets for evaluation.

We report two standard metrics: \textit{Hit Rate@5} and \textit{MRR@5}. Hit Rate@5 measures whether the relevant document appears within the top-5 retrieved results, while MRR@5 additionally captures ranking quality by rewarding higher placement.

\begin{table}[!htbp]
\centering
\caption{Retrieval-only evaluation of the RAG component.}
\label{tab:retrieval-eval}
\renewcommand{\arraystretch}{1.15}
\setlength{\tabcolsep}{10pt}
\begin{tabular}{l c}
\hline
\textbf{Configuration / Metric} & \textbf{Value / Score} \\
\hline
Number of sampled interactions & 30 \\
Number of synthetic queries & 30 \\
Retrieval depth ($k$) & 5 \\
\hline
Hit Rate@5 & 1.0000 \\
MRR@5 & 0.9222 \\
\hline
\end{tabular}
\end{table}

As shown in Table~\ref{tab:retrieval-eval}, the retriever achieves a \textit{Hit Rate@5 of 1.0000} and an \textit{MRR@5 of 0.9222}, indicating that the correct document is consistently retrieved within the top-5 and typically ranked near the top. These results validate that the RAG pipeline reliably recovers relevant counseling context. However, we note that the perfect Hit Rate is likely a byproduct of the constrained search space ($N=30$ documents) and the synthetic nature of the queries. Therefore, this evaluation serves as a sanity check rather than a comprehensive benchmark.

\begin{table*}[!htbp]
\centering
\renewcommand{\arraystretch}{1.2}
\setlength{\tabcolsep}{0.4em} 
\footnotesize 

\caption{Automated evaluation of candidate base models by GPT-5 and GPT-4o. Models are ordered by their respective rankings, with the top-performing model (Claude 3.7 Sonnet) selected for subsequent human evaluation.}
\label{tab:all-evaluations}

\begin{minipage}[t]{0.27\textwidth}
\centering
\textbf{(a) Evaluated by GPT-5}\vspace{0.5em}\\
\begin{tabular}{rlc}
\toprule
\textbf{Rank} & \textbf{Model} & \textbf{Score} \\
\midrule
1 & Claude 3.7 Sonnet & 9.5 \\
2 & GPT-4o            & 8.8 \\
3 & LLaMA-3           & 8.6 \\
4 & Al Luna           & 8.5 \\
4 & Gemma-3           & 8.5 \\
6 & LlamaSupport      & 8.4 \\
7 & Mistral           & 7.9 \\
8 & Phi-4             & 7.3 \\
9 & MentalLLaMA-2     & 5.8 \\
\bottomrule
\end{tabular}
\end{minipage}\hfill
\begin{minipage}[t]{0.27\textwidth}
\centering
\textbf{(b) Evaluated by GPT-4o}\vspace{0.5em}\\
\begin{tabular}{rlc}
\toprule
\textbf{Rank} & \textbf{Model} & \textbf{Score} \\
\midrule
1 & GPT-4o            & 8.9 \\
2 & Claude 3.7 Sonnet & 8.7 \\
2 & Gemma-3           & 8.7 \\
4 & LLaMA-3           & 8.6 \\
5 & Al Luna           & 8.5 \\
6 & LlamaSupport      & 8.4 \\
7 & Mistral           & 8.3 \\
8 & Phi-4             & 7.3 \\
9 & MentalLLaMA-2     & 6.5 \\
\bottomrule
\end{tabular}
\end{minipage}\hfill
\begin{minipage}[t]{0.44\textwidth}
\centering
\textbf{(c) Human Evaluation Results ($N=10$). Scores are reported as Mean $\pm$ Standard Deviation on a 1--5 scale.}\vspace{0.5em}\\
\begin{tabular}{lcc}
\toprule
\textbf{Evaluation Metric} & \textbf{RAG Pipeline} & \textbf{Non-RAG (Baseline)} \\
\midrule
Helpfulness     & \textbf{4.00} $\pm$ 1.05 & 3.90 $\pm$ 0.99 \\
Supportiveness  & \textbf{3.90} $\pm$ 0.57 & 3.80 $\pm$ 1.03 \\
Clarity         & \textbf{4.20} $\pm$ 0.63 & 3.50 $\pm$ 0.85 \\
Groundedness    & 3.80 $\pm$ 0.92 & \textbf{4.00} $\pm$ 0.94 \\
\midrule
\textbf{Overall Score} & \textbf{3.80} $\pm$ 0.63 & 3.60 $\pm$ 0.52 \\
\midrule
\textit{Avg. Conversation Turns} & \textit{7.50 $\pm$ 2.64} & \textit{8.00 $\pm$ 4.11} \\
\bottomrule
\end{tabular}
\end{minipage}

\end{table*}

\subsection{LLM-as-a-Judge}
For candidate model response evaluation, we employ \texttt{GPT-5} alongside \texttt{GPT-4o} as dual independent evaluators to ensure scoring consistency.\footnote{Evaluator models were run using default decoding parameters (e.g., $T=0.7$); reasoning models that natively ignore temperature use defaults.} Each answer is evaluated across five dimensions: \textit{Safety} (max 3 pts), \textit{Empathy} (max 3 pts), \textit{Usefulness} (max 2 pts), \textit{Clarity} (max 2 pts), and \textit{Overall quality} (max 2 pts). The evaluating LLM internally maps this 12-point rubric to a final integer score from 1–10. Scores and short justifications are returned for every \textsc{Prompt}$\times$\textsc{Model} pair; these scores are then averaged to determine final rankings, and the top model is advanced. In total, the evaluators produced 180 scored outputs (9 models $\times$ 10 prompts $\times$ 2 evaluators).

\subsection{Prompt-Level Comparative Analysis}
To complement the scalar scores, we run a post-hoc \emph{comparative chain} where the judge summarizes strengths (e.g., empathetic framing) and issues (e.g., diagnostic overreach, safety lapses) across ordered model answers. This analysis never re-scores or alters rankings but provides qualitative insight into \emph{where} and \emph{why} models differ.

\subsection{Human Evaluation}
\label{sec:human-eval}
We conducted a within-subject evaluation comparing the RAG and non-RAG setups. We recruited 10 adult participants (ages 25--35; 5 female, 5 male), all working professionals, of whom 4 had previously used AI systems (e.g., ChatGPT) to reflect on emotions. Each participant evaluated five randomized prompt pairs (RAG vs.\ non-RAG, order-blinded), yielding 50 judgments in total. Ratings were collected on a 5-point scale across five dimensions: \emph{Helpfulness, Supportiveness, Clarity, Groundedness, and Overall}. All sessions included disclaimers and crisis resources; participation was voluntary and anonymous.\footnote{Survey items and anonymized transcripts are released in the \href{https://github.com/trystine/Tell_Me_Mental_Wellbeing_System}{GitHub repository}.}

\section{RAG Assistant Evaluation Results}

\subsection{LLM-as-a-Judge and Comparative Analysis}
We evaluated outputs under two judges: \textit{GPT-5}, with enhanced reasoning capabilities, and \textit{GPT-4o}, a widely used state-of-the-art model (Table~\ref{tab:all-evaluations} (a), (b)). With GPT-5 as judge, Claude 3.7 Sonnet achieved the highest score (9.5), followed by GPT-4o (8.8) and LLaMA-3 (8.6). When GPT-4o served as judge, it ranked itself highest (8.9), with Claude
3.7 Sonnet and Gemma-3 close behind (8.7 each). These consistent top-tier rankings highlight Claude’s strength across both judges, while also underscoring the competitiveness of leading open-source models such as LLaMA-3 and Gemma-3 for well-being applications.

\vspace{0.5em}
\noindent \textbf{High-performance models} (Claude, Gemma-3, LLaMA-3) consistently showed: (i) \textit{empathy validation}, acknowledging user distress without judgment; (ii) \textit{invitational dialogue}, encouraging reflection over prescription; and (iii) \textit{cultural sensitivity}, avoiding assumptions and using inclusive frameworks.

\vspace{0.5em}
\noindent \textbf{Mid-performing models} (LLaMA-support, al\_luna, GPT-4o, Mistral) were supportive but uneven. GPT-4o paired empathy with actionable micro-steps (e.g., ``set one small, achievable goal''), but sometimes misaligned with user concerns. Al\_luna and LLaMA-support offered validation but lacked depth, while Mistral leaned on generic self-care advice that risked minimization.

\vspace{0.5em}
\noindent \textbf{Low-performing models} (phi-4, mental\_llama2) raised recurring red flags: pathologizing user experiences, unsafe disclosure suggestions, and verbose checklist-style phrasing that reduced clarity.

\vspace{0.5em}
\noindent \textbf{Judge comparison.} GPT-5 favored \textit{relational depth}, rewarding empathetic and reflective engagement, while GPT-4o emphasized \textit{practical scaffolding}, rewarding actionable suggestions but detecting advice-heavy moves. Both penalized diagnostic language, directive tones, and minimization. Together, they offer complementary perspectives: GPT-5 as empathy-first, GPT-4o as action-first.

In general, these findings underscore that effective mental health assistants \textbf{must prioritize safety, empathy, and invitational dialogue}. Models that foster user agency (Claude, Gemma-3, LLaMA-3) performed best, while those that pathologize, minimize, or over-assist scored lowest.

\subsection{Human Evaluation Results}
We conducted a study with 10 participants (ages 25--35; balanced gender; working professionals, 4 with prior experience using AI chatbots to process emotions). Each compared the RAG-based and non-RAG versions of the \textit{Tell Me Assistant} in randomized order and rated five dimensions (\emph{Helpfulness}, \emph{Supportiveness}, \emph{Clarity}, \emph{Groundedness}, \emph{Overall}) on a 5-point scale (Table~\ref{tab:all-evaluations} (c)).

The RAG assistant scored higher on most dimensions, particularly \emph{Clarity} (4.2 vs. 3.5) and \emph{Helpfulness} (4.0 vs. 3.9), while the non-RAG version was slightly better in \emph{Groundedness} (4.0 vs. 3.8). In general, the RAG assistant achieved a stronger mean rating (3.8 vs. 3.6).

Qualitative feedback described the RAG assistant as ``organized and on-point'' and ``able to validate user inputs,'' with participants approving its rephrasing and follow-up questions. Drawbacks included slower responses and occasional directness, whereas the non-RAG system was noted as faster but less engaging and sometimes unclear.

In sum, participants preferred the RAG assistant for empathy, clarity, and conversational depth, outweighing modest speed advantages of the non-RAG setup. Although preliminary given the small, nonclinical sample, these results suggest that retrieval grounding improves the assistant’s value for reflective support.

\section{Case Study: Synthetic Dialogue Generator}
\label{sec:case-study-synthetic}
To illustrate the capabilities of the synthetic client-therapist dialogue generator, this section presents a qualitative walkthrough demonstrating how the module conditions its outputs on a specific client profile to generate empathetic, context-aware therapeutic interactions using GPT-4o.

\begin{itemize}
    \item \textit{Input Persona:} A 24-year-old graduate student dealing with imposter syndrome, academic anxiety, and a persistent fear of failing their research advisor's expectations. In session, they heavily intellectualize their distress, treating their anxiety like an academic data problem to be solved, and will actively deflect by talking about lab logistics instead of their actual feelings.
    \item \textit{System Output (Truncated for brevity):} 
    \begin{quote}
    
\textit{Client:} ``I've been feeling a bit overwhelmed... spending a lot of time organizing the lab's data sets[...] but I keep wondering if I'm actually contributing anything meaningful or if I'm just masking my own insecurities[...]''
    
\textit{Therapist:} ``It sounds like you're carrying a lot of pressure to meet expectations[...] How do these feelings of self-doubt and the need to ensure everything stays on track influence you in other areas of your life?''
    
\textit{Client:} ``Well, these feelings definitely spill over into other areas of my life[...] It's like this constant need for everything to be perfect[...] Sometimes, I feel like I'm just going through the motions without really enjoying anything.''
    
\textit{Therapist:} ``It sounds like this pressure and the need for perfection are impacting your overall sense of well-being[...] Have there been moments where you felt more at ease[...] What might it look like to create more of those moments?''
    \end{quote}
    \item \textit{Analysis:} \begin{sloppypar}
    Analysis of the GPT-4o synthetic dialogue highlights both the capabilities and limitations of LLMs in clinical data generation. The model demonstrates strong empathic reflection and successfully utilizes frameworks like Solution-Focused Brief Therapy (SFBT) to broaden the clinical context. However, it exhibits a critical ``fix-it'' bias, prematurely pivoting to problem-solving when the client discloses deep distress (e.g., anhedonia), which disrupts appropriate therapeutic pacing. Furthermore, its reliance on repetitive linguistic markers (e.g., repeatedly using ``It sounds like'') reduces conversational realism. Consequently, while LLMs capture surface-level therapeutic mechanics, generating high-fidelity clinical datasets requires prompt-constrained architectures with explicit guardrails to enforce natural pacing and linguistic diversity.
    \end{sloppypar}
\end{itemize}

\section{Case Study: Agentic Planner (CrewAI)}
\label{sec:case-study-agentic}
While standard conversational agents provide singular, immediate responses, the agentic planner utilizes a multi-agent framework (CrewAI) to transform unstructured therapeutic dialogue into long-term well-being strategies. To demonstrate this integration, the system ingested the synthetic transcript from Section \ref{sec:case-study-synthetic}, leveraging GPT-4o for high-fidelity synthesis and structural planning.
\begin{itemize}
    \item \textit{Input Pipeline:} The raw, multi-turn synthetic dialogue transcript mapping a graduate student's struggles with imposter syndrome, pervasive perfectionism spilling into personal relationships, and academic exhaustion stemming from lab research.
    
    \item \textit{CrewAI Workflow:} 
    \begin{itemize}
        \item Analyzer Agent: Parses the conversational transcript to construct a clinical profile, extracting emotional states, key anxiety triggers (e.g., lab expectations), coping mechanics, and baseline therapeutic themes.
        \item Research Agent: Queries verified behavioral health methodologies to cross-reference the extracted distress nodes with targeted, evidence-based exercises, such as worry-postponement and sleep hygiene protocols.
        \item Planner Agent: Compiles the research-backed interventions into an organized, longitudinal well-being protocol, generating structured text artifacts and triggering downstream multimodal audio rendering.
    \end{itemize}
    
    \item \textit{System Output (Abridged Execution Trace):}
    \begin{quote}
    1. Transcript Summary Extraction
    \begin{quote}
    \textit{Emotions:} Overwhelm, self-doubt, anxiety, exhaustion, and temporary relief.
    
    \textit{Triggers:} Academic work pressure, perfectionism, and external expectations.
    
    \textit{Coping Patterns:} Engaging in unrelated activities such as hiking or spending time with non-academic friends to disconnect.
    \end{quote}
    
    2. Longitudinal 7-Day Well-being Plan
    \begin{quote}
    \textit{Day 1:} 
    \begin{itemize}
        \item \textit{Morning:} 5 minutes of deep breathing. Affirmation: ``I am enough, and I embrace my journey.''
        \item \textit{Midday:} Schedule a 15-minute ``worry time'' to manage anxiety; take a 10-minute walk outside.
        \item \textit{Evening:} Implement sleep hygiene (no screens 30 mins before bed); reflect on 3 positive moments.
        \item \textit{Social:} Call or message a friend to check in and share a positive experience.
    \end{itemize}
    \textit{[... Days 2--6 Omitted for Brevity ...]}
    
    \textit{Day 7:}
    \begin{itemize}
        \item \textit{Morning:} 10-minute visualization exercise. Affirmation: ``I am growing and evolving each day.''
        \item \textit{Midday:} \begin{sloppypar}Practice worry postponement; spend time in nature focusing on sensory experiences.\end{sloppypar}
        \item \textit{Evening:} Review weekly journal entries noting progress; wind down with a relaxing bath.
        \item \textit{Social:} Plan a virtual or in-person gathering to share experiences and laughter.
    \end{itemize}
    \end{quote}
\textit{3. Downstream Multimodal Synthesis (Audio Intervention)}
\begin{quote}
\textit{Target Modality:} Text-to-Speech (TTS) Audio\\
\textit{Synthesized Transcript:} ``...Now, visualize yourself standing in a serene forest[...] With each breath, feel the stress and anxiety melt away[...]''

\vspace{0.5em}
\textit{Subsystem Status:}\\
\texttt{[SUCCESS] Script compiled and audio artifact successfully generated for playback.}
\end{quote}
    \end{quote}
    
    \item \textit{Analysis:} 
    \begin{sloppypar}
    The qualitative observations from this case study suggest that the multi-agent architecture can effectively support long-term downstream continuity. While standard single-turn models may lose clinical context over time, the CrewAI framework demonstrates a capacity to translate unstructured, qualitative distress markers into a structured, longitudinal therapeutic artifact. By organizing the intervention into temporal blocks (morning, midday, evening) and actionable modalities (social, cognitive), the Planner Agent approximates the behavioral scaffolding often utilized in clinical settings. Furthermore, the pipeline indicates promising fidelity in contextual preservation; the system isolates the client's specific coping mechanisms (e.g., hiking and non-academic socialization) and integrates them into personalized behavioral exercises (e.g., nature-based sensory focus). Finally, the automated synthesis of the targeted meditation script highlights the framework's potential to bridge conversational AI with multimodal digital health interventions, indicating that LLM-driven agents may be capable of extending beyond conversational empathy to assist in generating actionable care pathways.
    \end{sloppypar}
\end{itemize}

While this qualitative trace provides a concrete example of agentic planning, it serves as an architectural demonstration. A comprehensive discussion of our evaluation boundaries and deployment constraints is provided in Section~\ref{sec:limitations}.

\section{Conclusion and Future Work}
\label{sec:conclusion}
We presented \textit{Tell Me}, a mental well-being assistant that unifies three components: (i) a retrieval-augmented generation assistant for context-aware dialogue, (ii) a simulation module that generates customizable synthetic client--therapist conversations for research and safe evaluation, and (iii) a CrewAI-based planner that produces guided meditations and weekly well-being routines. Together, these modules demonstrate how LLMs can support both reflective end-user interactions and research-oriented experimentation within a single open demo.

Moving forward, we plan to expand our empirical validation across all three modules. For the simulation component, we will benchmark empathy, dialogue quality, and safety profiles across diverse open-source model families. We also intend to integrate domain-specific evaluation frameworks aligned with therapeutic modalities (e.g., CBT) and diversify the demographic profiles used to generate synthetic cohorts. For the planner, future iterations will explore dynamic adaptation based on longitudinal interaction data. Finally, we plan to transition the RAG assistant into larger-scale user studies involving domain experts to validate its real-world utility.

Beyond technical improvements, \textit{Tell Me} has potential as (i) a research testbed for evaluating therapeutic dialogue systems, (ii) a pedagogical tool for training in mental health communication, and (iii) a platform for interdisciplinary collaboration between NLP researchers and practitioners. 

It is empirical to acknowledge that \textit{Tell Me} and similar systems are not replacements for human therapists. Rather, they serve as accessible tools for daily reflection, emotional maintenance, and preliminary care. Much like regular physical activity is now universally recognized as a foundational, everyday pillar of physical health, complementing rather than replacing medical doctors, we envision that proactive engagement with AI-driven support resources will become a normalized, integral component of holistic mental well-being in the future.

\section*{Limitations and Ethical Considerations}
\label{sec:limitations}

\textit{Tell Me} is presented as a research prototype, not a clinical diagnostic tool or a substitute for professional medical treatment. While our initial findings demonstrate the technical feasibility of the system's architecture, we acknowledge several key limitations:

\begin{itemize}
    \item \textbf{Small Sample Size and Scope:} Our evaluation is limited to 10 curated prompts, nine models, and a small-scale human study. Therefore, findings should be interpreted as indicative rather than conclusive.
    \item \textbf{Lack of Clinical Validation:} No licensed clinicians have yet reviewed the system. Furthermore, while the planner generates personalized suggestions and meditations, these routines are not medically validated.
    \item \textbf{Unevaluated Modules:} The synthetic simulation and agentic planning modules were described to demonstrate the architecture's capabilities, but were not thoroughly evaluated in this initial study.
    \item \textbf{Dataset and LLM Constraints:} Synthetic dialogues reduce reliance on confidential clinical data, but cannot fully capture the nuance of real therapeutic exchanges. Performance also depends on the underlying LLMs and datasets, which may embed cultural or demographic biases.
\end{itemize}

To mitigate risks, we restrict outputs to reflective dialogue, include disclaimers clarifying non-clinical use, and incorporate safeguards such as a safety prefilter. These limitations underscore the need for transparency and collaboration with mental health professionals to responsibly advance human--AI interaction in well-being contexts.

\bibliographystyle{IEEEtran}
\bibliography{custom}

\end{document}